# Movement Optimization of Robotic Arms for Energy and Time Reduction using Evolutionary Algorithms

Abolfazl Akbari[1], Saeed Mozaffari[2], Rajmeet Singh[2], Majid Ahmadi[1], Shahpour Alirezaee[2]
[1]Electrical and Computer Engineering Department. University of Windsor, Windsor, Canada
[2]Mechanical, Automotive, and Material Engineering Department. University of Windsor, Windsor, Canada
{akbari41, saeed.mozaffari, rsbhourji, m.ahmadi, s.alirezaee } @uwindsor.ca

*Abstract*— Trajectory optimization of a robot manipulator consists of both optimization of the robot movement as well as optimization of the robot end-effector path. This paper aims to find optimum movement parameters including movement type, speed, and acceleration to minimize robot energy. Trajectory optimization by minimizing the energy would increase longevity of robotic manipulators. We utilized particle swarm optimization method to find the movement parameters leading to minimum energy consumption. The effectiveness of the proposed method is demonstrated on different trajectories. Experimental results show that 49% efficiency was obtained using a UR5 robotic arm.

## I. Introduction

The field of robotics has been rapidly advancing over the past few decades. The use of robotic arms is becoming increasingly prevalent in various applications, ranging from manufacturing and industrial automation to healthcare and space exploration. Robotic arms are often used for tasks that require repetitive movements or heavy lifting, such as assembly line operations in manufacturing. Trajectory is defined is defined as the robot position as a function of time. In other words, it is a combination of path and time scaling. While path scaling is described as geometric description of the end-effector locations, time scaling specifies the time required for moving from one position to another.

Several approaches have been proposed for optimizing the trajectory of a robot. Optimization approaches aim to minimize/maximize at least one of the following objective functions [1]: minimum time trajectory planning [2]; minimum energy trajectory planning [2]; minimum jerk trajectory planning [3]. Minimum time trajectory planning was the first criterion as it directly relates to reducing manufacturing time and increasing productivity. Minimizing robot energy leads to mechanical stresses reduction in actuators and energy cost. Trajectory planning is based on minimizing Jerk, the third derivative of the position in time, reduces joint positioning errors and robot vibrations. These objectives can be considered simultaneously to achieve better results. For example, optimal time–jerk trajectory planning was performed in [4] by improved butterfly optimization algorithm (IBOA) for Delta parallel robot. Minimum time-energy path planning was suggested for multi-robots to avoid collision in an unknown environment [5].

The high energy consumption associated with robotic arms remains a significant challenge. The largest proportion of energy usage in a car manufacturing facility, for example, is attributed to industrial robots, which operate using mechanical components that move. To address this issue, a lot of industrial companies endeavor to minimize their robot's energy consumption. The trajectory optimization for a robotic arm by minimizing the energy has several advantages. By minimizing the energy required for the movement of the robotic arm, this technique ensures that the system operates efficiently and effectively. Moreover, it reduces the wear and tear of the system and increases its lifespan. Additionally, it leads to reduced energy consumption, which is particularly important for applications where the power supply is limited.

In this paper, we are interested in minimum energy trajectory planning of a robotic arm as well as operation time by using particle swarm optimization (PSO). We assume that the robot trajectory is completely specified by the task and the end-effector is required to pass through known positions, called waypoints. For example, in a pick-and-place task, the pick and place locations are determined by the application. Therefore, unlike previous efforts, we mainly focus on the time scaling rather than path generation. We aim to find robot movement parameters with the PSO algorithm to minimize energy consumption and satisfy the robot velocity and acceleration constraints.

## II. Trapezoidal Motion Profile

We assume point-to-point trajectory which involves starting from a position and coming to a stop in a different position. These positions are called waypoints in robotics. Waypoints are set by operators to define the trajectory of a robot's movement or to specify a specific path that the robot should follow to complete a task.

### A. Movement Types

There are two types of movements for point-to-point motion. The simplest type is move in a straight line from one waypoint to another in a continuous motion. This movement is called *MoveL* (Linear motion), and the robot will follow a straight-line path between waypoints while maintaining a constant orientation. This type of movement is used when the path of robot's end-effector is important, and the robot must work in a confined space. MoveP (Process motion) is similar to MoveL but the robot's end-effector maintains the same speed through several waypoints. This type of motion is used in process applications such as welding or gluing in which the quality of the process depends on the robot's speed. While robot works in free space, it can follow a non-linear path that is defined by its joint angles. This type of movement is called *MoveJ* (Joint motion) and the robot will move its joints in a coordinated motion to reach the specified position.

### B. Time Scaling

Trapezoidal velocity profile is the most common approach for industrial robot's trajectory planning [6]. The operator determines the maximum velocity and acceleration for the motion, which in turn determines the trapezoidal velocity profile of the robot. Figure 1 shows the Universal robot interface through which the motion parameters are set by an operator.

Trapezoidal time scaling gets its name from their velocity profiles (Figure 2). It consists of a constant acceleration phase $\ddot{s}(t) = a$ of time $t_a$, followed by a constant velocity phase $\dot{s}(t) = v$ of time $t_v = T - 2t_a$, followed by a constant deceleration phase $\ddot{s}(t) = -a$ of time $t_a$. It is shown that the trapezoidal time scaling is the fastest straight-line motion when there are known constant limits on the joint velocities and accelerations [7].

To describe this motion profile only two parameters $(a, v, T)$ can be selected, because there is a dependency between them, and they must satisfy $S(T) = 1$ and $v = a \times t_a$. By substitution $t_a = v/a$, we can describe the motion profile during the three stages as follows:

$$\text{for} \quad 0 \leq t \leq \frac{v}{a} \quad \begin{aligned} \ddot{s}(t) &= 0 \\ \dot{s}(t) &= at \\ s(t) &= \frac{1}{2}at \end{aligned} \quad (1)$$

$$\text{for} \quad \frac{v}{a} \leq t \leq T - \frac{v}{a} \quad \begin{aligned} \ddot{s}(t) &= 0 \\ \dot{s}(t) &= v \\ s(t) &= vt - \frac{v^2}{2a} \end{aligned} \quad (2)$$

$$\text{for} \quad T - \frac{v}{a} \leq t \leq T \quad \begin{aligned} \ddot{s}(t) &= -a \\ \dot{s}(t) &= v \\ s(t) &= \frac{2avT - 2v^2 - a^2(t-T)^2}{2a} \end{aligned} \quad (3)$$

## III. PROPOSED METHOD

For designing an energy-efficient trajectory, appropriate control actions to all joint actuators (movement type, speed, acceleration) should be chosen to force the manipulator's end-effector to follow a time-bounded predefined trajectory. In this paper, the optimum movement parameters are selected by an evolutionary algorithm.

### A. Optimization constraints

Since only two of $v$, $a$, and $T$ can be chosen independently, we have some constraints.

(i): If $v$ and $a$ are specified, then $v^2/a < 1$ should be satisfied to reach the maximum velocity $v$. Otherwise, the trapezoidal profile will change to a triangular profile with an acceleration to velocity $v$ in time $t_a = v/a$ followed by an immediate deceleration to rest in time $t_a$. The area of this profile is $2t_a \times v/2 = v^2/a$. To satisfy constraint $S(T) = 1$, $v^2/a$ should be less than 1. The minimum possible time for the motion can be obtained from equation (3):

$$T = \frac{a + v^2}{va} \quad (4)$$

(ii): If we choose $v$ and $T$ such that $1 < vT < 2$, ensuring a trapezoidal profile and that the top speed $v$ is sufficient to reach $s = 1$ in time $T$, and solve $S(T) = 1$ we obtain:

$$a = \frac{v^2}{vT - 1} \quad (5)$$

(iii): If $a$ and $T$ are selected such that $aT^2 > 4$, ensuring that the motion is completed in time, and solve $S(T) = 1$ then:

$$v = \frac{1}{2}(aT - \sqrt{a}(\sqrt{aT^2 - 4})) \quad (6)$$

### B. Optimization Criterion

In this study, we choose two main objective functions which are Cycle time and Energy consumption. Energy consumption optimization and cycle time optimization are two important objectives in manufacturing processes, and they are often inversely related in practice.

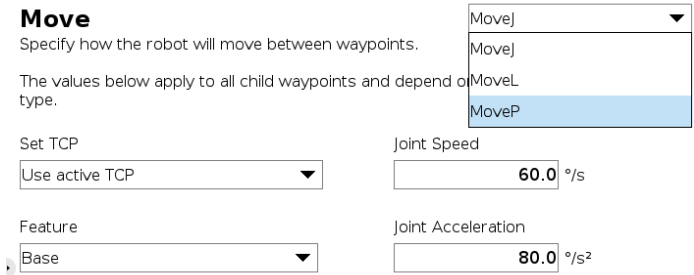

Figure 1. *Movement parameters in Universal robots.*

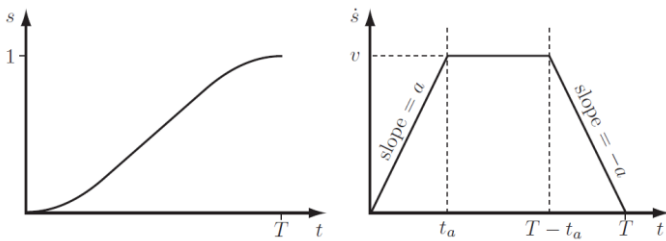

Figure 2. *Time scaling and its derivative for a trapezoidal motion profile.*

It should be noted that reducing cycle time generally involves increasing the speed of the manufacturing process, which can lead to higher energy consumption. On the other hand, reducing energy consumption may require slowing down the manufacturing process, which can increase cycle time.

We consider the overall energy consumption of all six joints of the robotic arm as the objective function:

$$S_1 = \sum_{m=1}^{M} \sqrt{\frac{1}{T_i} \int_0^{T_i} (a(t))^2 dt} \, T_i \quad (7)$$

In this equation, $T_i$ is the total operation time, $a(t)$ is the acceleration curve of each joint, and $m = 1, 2, ..., M$ is the number of joints.

Cycle time optimization means that when the kinematic and dynamic constraints are met, the manipulator completes the specified task in the shortest time to enhance the work efficiency of the system. To improve the operation efficiency of the robot, the operation time of the robot should be minimized for the robot to move along the specified trajectory under the condition of meeting various constraints and it can be defined as:

$$S_2 = \sum_{i=1}^{n-1} (t_{i+1} - t_i) \quad (8)$$

In this formula, $n$ is the number of waypoints.

Since there is a trade-off between energy consumption and cycle time, the chosen fitness function (optimization criterion) $f_f$ places the same weight on both these factors:

$$f_f = \frac{S_1}{2} + \frac{S_2}{2} \quad (9)$$

### C. Optimization Process

The proposed energy minimization method is based on the PSO algorithm to calculate the optimum values of $v$ and $a$ for each point-to-point trajectory, considering the constraints mentioned before. We use dynamic model of a 6 DoF manipulator in MATLAB to perform the movement of a robot manipulator with the obtained movement parameters.

Particle swarm optimization is a metaheuristic optimization technique that is inspired by the social behavior of swarms [8]. We selected this optimization algorithm because of its simplicity and ease of implementation. It also has a fast convergence rate and can handle large-scale optimization problems.

The algorithm starts with a swarm of particles that move around a search space to find the optimal solution. Each particle represents a potential solution, and its position in the search space corresponds to a set of values for the variables of the optimization problem.

At each iteration of the algorithm, each particle evaluates its current position ($x_{ij}^t$) according to the objective function (Equation 2) and updates its velocity ($v_{ij}^t$) and position based on its personal best solution ($l_{ij}^t$) and the best solution found by its neighbors ($g_j^t$). The objective of PSO is to find the best solution by updating the position of each particle based on its own experience and the experience of its neighboring particles.

$$v_{ij}^{t+1} = w \times v_{ij}^t + c_1 r_1 [l_{ij}^t - x_{ij}^t] + c_2 r_2 [g_j^t - x_{ij}^t] \quad (10)$$

where $v_{ij}^t$ is the velocity of particle $i$ in dimension $j$ at time $t$ and $w$ is the inertial weights. Parameters $c_1$ and $c_2$ are constant values and $r_1$ and $r_2$ are uniform random numbers. Positions are calculated as the addition of the actual velocity to the previous position:

$$X_i^{t+1} = X_i^t + V_i^{t+1} \quad (11)$$

At the end, we calculate fitness function $f_f$ value for each value of $v$ to find the lowest fitness function value.

## IV. EXPERIMENTAL RESULTS

This section explains the utilized robotic arm, tested movements, and the impact of the proposed method on the robot's energy reduction.

### A. Setup

The proposed method was evaluated on MATLAB with the obtained movement parameters by the PSO algorithm. The optimum movement type, speed, and acceleration were demonstrated and verified on a real-world robot manipulator. We used Universal Robot UR5 which is a 6 DoF collaborative robot with payload of 5 kg and 850 mm reach.

We have chosen two trajectories for the end-effector movements to investigate the effectiveness of the proposed energy saving method. We performed our experiments by using the *MoveJ* command, which produces linear curves in the joint space blended by a set acceleration of nearby endpoints. Figure 3 shows the tested movements which include linear segments of different lengths and various direction changes.

### B. Results

By applying the constraints that explained in previous section, we expected that our model produces same velocity profile as shown in figure 2. Figure 4 shows the velocity profile of the robot for moving from one point to another that is similar to the theoretical velocity profile described in figure 2. Figure 5 demonstrates normalized values of cycle time and energy consumption as a function of end-effector velocity.

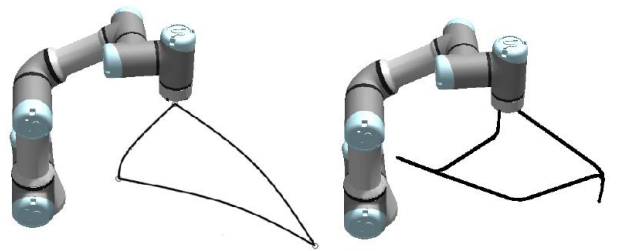

Figure 3. *Visualization of the two testing robot end-point paths; the UR robot is shown as scale reference here.*

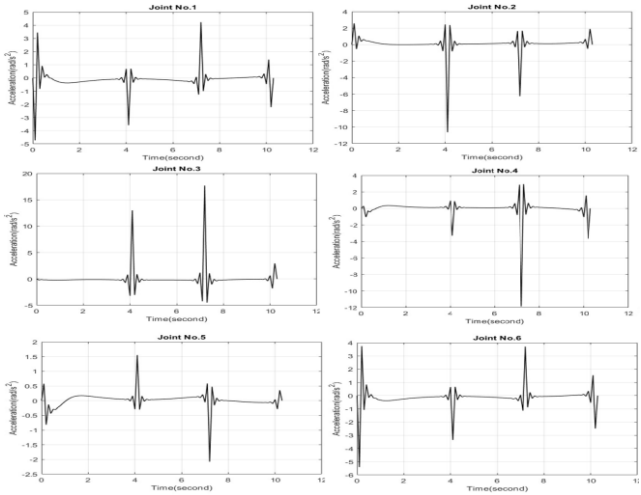

Figure 4. *Joint acceleration at lowest value of the fitness function.*

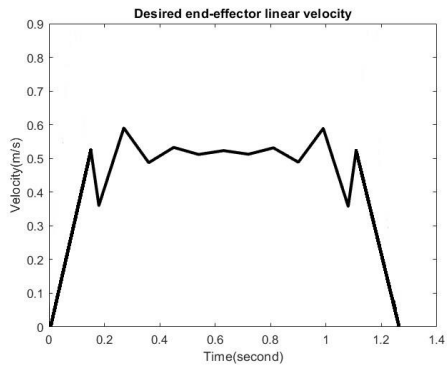

Figure 5. *Trapezoidal velocity profile.*

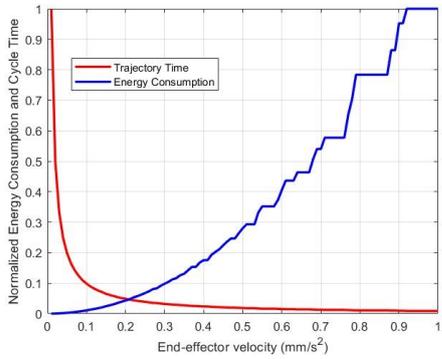

Figure 6. *Normalized values of Energy consumption and Trajectory time against velocity of the end-effector.*

For each trajectory, the values of $f_f$ were calculated based on the MATLAB simulation for different values of end-effector velocity. Experiments show the optimal end-effector velocity of 0.18 mm/s$^2$ which is close to the intersection of objective functions (0.2 mm/s$^2$). The lowest, worst and the average values of $f_f$ are shown in Table 1 which indicate the percentage of the fitness function:

$$F = 100 \times (1 - \frac{f_f^B}{f_f^i}) \qquad (12)$$

Table 1. Values of fitness functions and percentage of improvement

| \multicolumn{5}{c}{Results of Fitness function & Improvement} |
|---|---|---|---|---|
| $f_f^{best}$ | $f_f^{average}$ | $f_f^{worst}$ | $F^{worst}$ | $F^{average}$ |
| 0.0443 | 0.0869 | 0.2119 | 78.22% | 49.09% |

In which $f_f^i$ is the value of fitness function at each velocity and $f_f^B$ is the lowest value of fitness function.

In this table, the lowest value of fitness function is 0.0443 which corresponds to certain value of velocity of the end-effector that minimizes both defined objective functions (7) (8). The percentage of improvement that the best velocity compared to the worst value of velocity is around 78% which theoretically represents the greatest possible improvement in reducing energy consumption and operation cycle time. If we consider a random velocity value versus the best value, the percentage of average improvement would be 49%.

## V. CONCLUSION

The goal of this paper was to present a methodology for movement optimization of a universal robot to optimize energy consumption of the robot as well as reducing operation time of the robot for a given trajectory of movement of the manipulator endpoint. The optimization algorithm was based on the PSO algorithm. The results show that if we chose the best value of velocity of end-effector rather than a random velocity, the percentage of improvement would be 49%.